\documentclass[unnumsec,webpdf,contemporary,large]{oup-authoring-template}%

\usepackage{microtype}
\usepackage{tabularx} 
\usepackage{subfigure}
\usepackage{makecell}
\usepackage{array}
\makeatletter
\renewcommand{\@thesubfigure}{\hskip\subfiglabelskip\footnotesize \thesubfigure \hskip\subfiglabelskip}
\makeatother

\graphicspath{{Fig/}}

\theoremstyle{thmstyleone}%
\theoremstyle{thmstyletwo}%
\theoremstyle{thmstylethree}%

\begin{document}

\journaltitle{}
\DOI{DOI HERE}
\copyrightyear{2022}
\pubyear{2019}
\access{Advance Access Publication Date: Day Month Year}
\appnotes{Paper}

\firstpage{1}


\title[Short Article Title]{PVminer: A Domain-Specific Tool to Detect the Patient Voice in Patient Generated Data}

\author[1,2$\ast$]{Samah Fodeh\ORCID{0000-0003-4664-3143}}
\author[1]{Linhai Ma\ORCID{0000-0001-8519-864X}}
\author[1]{Yan Wang}
\author[1]{Srivani Talakokkul}
\author[1]{Ganesh Puthiaraju}
\author[1]{Afshan Khan}
\author[3]{Ashley Hagaman}
\author[3]{Sarah Lowe}
\author[4] {Aimee Roundtree} 

\authormark{Fodeh et al.}

\address[1]{\orgdiv{Department Of Emergency Medicine}, \orgname{Yale School of Medicine}, \orgaddress{\street{464 Congress Ave}, \postcode{06519}, \state{CT}, \country{USA}}}

\address[2]{\orgdiv{Department of Biomedical Informatics \& Data Science}, \orgname{Yale School of Medicine}, \orgaddress{\street{100 College Street}, \postcode{06510}, \state{CT}, \country{USA}}}

\address[3]{\orgdiv{Department of Social and Behavioral Sciences}, \orgname{Yale School of Public Health}, \orgaddress{\street{60 College Street}, \postcode{06520}, \state{CT}, \country{USA}}}

\address[4]{\orgdiv{Division of Research}, \orgname{Texas State University}, \orgaddress{\street{601 University Dr.}, \postcode{78666}, \state{TX}, \country{USA}}}


\corresp[$\ast$]{Corresponding author. \href{email:email-id.com}{samah.fodeh@yale.edu}}

\abstract{
\textbf{Motivation:} Patient-generated text such as secure messages, surveys, and interviews contains rich expressions of the patient voice (PV), reflecting both communicative behaviors and social determinants of health (SDoH). Traditional qualitative coding frameworks are labor intensive and do not scale to the large number of patient-authored messages produced across health systems. Existing machine learning (ML) and natural language processing (NLP) approaches provide partial solutions for analyzing patient-generated text. However, most methods treat patient-centered communication (PCC) and social determinants of health (SDoH) as separate tasks, ignore overlapping expressions within the same message. Some of them rely on pre-trained models that are not well suited to patient-facing language. These limitations highlight the need for scalable, domain-adapted approaches that jointly model the linguistic and social dimensions of patient-generated communication.\\
\textbf{Method:} We introduce PVminer, a domain-adapted NLP framework for structuring patient voice in secure patient-provider communication. PVminer formalizes PV detection as a multi-label, multi-class prediction task and integrates three components: pre-training of patient-specific BERT encoders (PV-BERT-base and PV-BERT-large), unsupervised topic modeling for thematic augmentation (PV-Topic-BERT), and fine-tuned classifiers of Code, Subcode, and Combo-level labels. Topic information derived from a PV-Topic-BERT model are concatenated during both fine-tuning and inference to enrich input representations with high-level semantic cues.\\
\textbf{Results:} PVminer with our proposed domain-adapted pre-trained models achieves strongest performance across all three hierarchical tasks and outperforms biomedical and clinical pre-trained baselines. For Code-level classification,  PV-BERT-large attains an F1 score of 82.25 \%. At the Subcode level, PV-BERT-base achieves the highest F1 score of 80.14 \%. For the more challenging Combo-level task, PV-BERT-base and PV-BERT-large obtain F1 scores of 77.58 \% and 77.87 \%. An ablation study further shows that author identity and topic-based augmentation each contribute meaningful gains.\\
\textbf{Availability:} Pre-trained models, source code, and documentation will be publicly released. Annotated datasets will be made available upon request for research use.\\
}

\maketitle

\section{Introduction}\label{sec1}

\begin{figure*}[ht]
\centering
\includegraphics[width=0.85\textwidth]{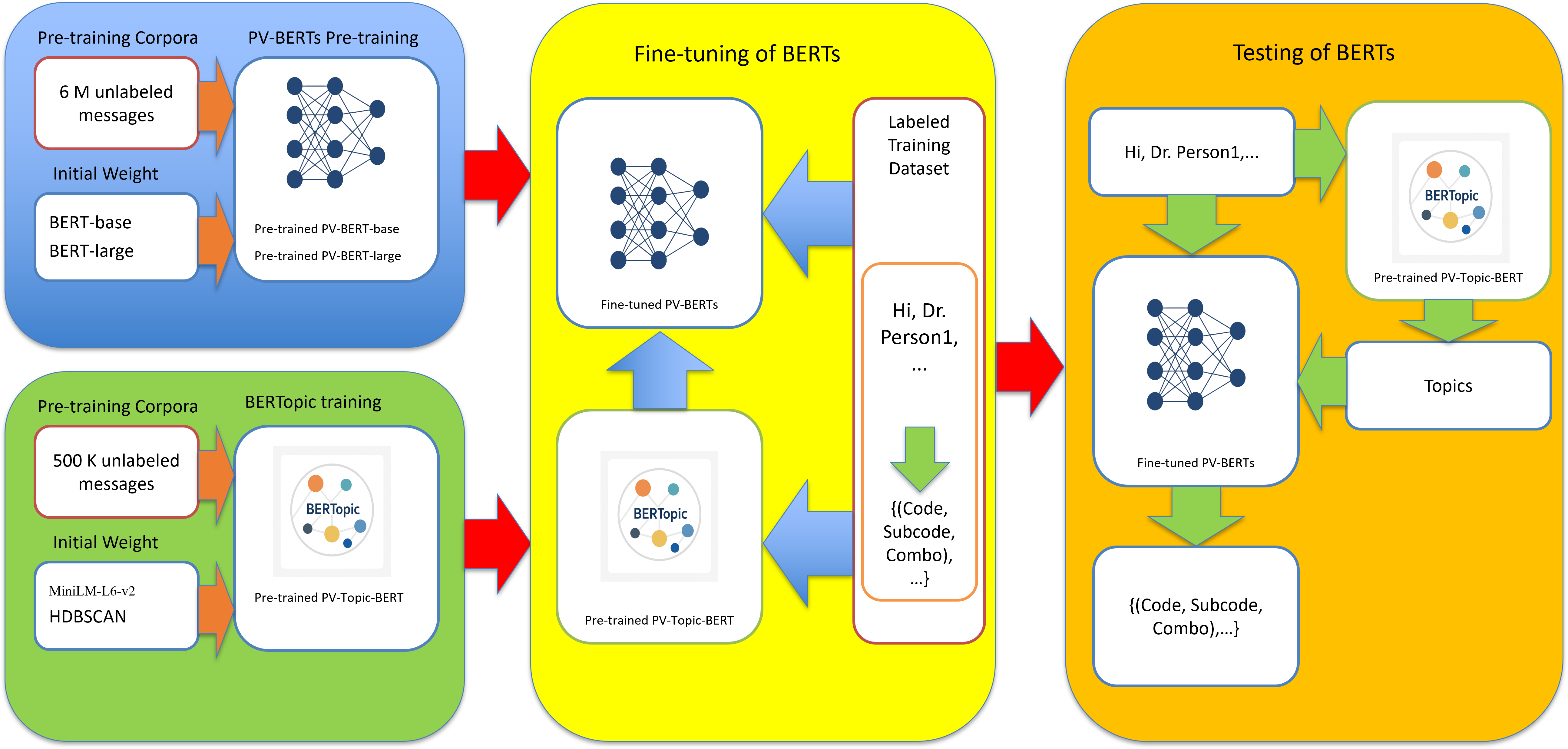}
\caption{
Overview of the PVminer framework. 
(1) A BERT baseline model is domain-adapted by pre-training on 6M unlabeled patient-authored messages using a masked language modeling objective, producing PV-BERT-base and PV-BERT-large. 
(2) A BERTopic model is trained on 500K unlabeled messages to learn latent thematic structure and generate topic keywords, producing PV-Topic-BERT.  
(3) During fine-tuning, PV-BERT models are augmented with topic cues from the pre-trained PV-Topic-BERT model and trained on annotated patient voice data to predict structured labels (Code, Subcode, Combo). 
(4) During inference, given a new message, topic cues are again retrieved from the PV-Topic-BERT model, and the fine-tuned PV-BERT produces the structured patient voice representation.
}\label{flowchart}
\end{figure*}

Patient-generated data are health relevant data captured by patients outside of traditional care settings through sensors or wireless devices, also collected in unstructured textual form through interview responses, open-ended questions on patient forms, or secure messages (SM) sent via patient portals. \cite{intro1, intro2, intro3, intro4} These text data capture the patients’ voice (PV) signifying their wants, needs, and fears that affect their outcomes. The PV not only represents the living experiences of patients related to environmental, social, and clinical domains (i.e., Social determinants of Health (SDoH)) but also the patients’ communication (PC) with providers representing their engagement and preferences regarding plan of care (e.g., shared-decision making and partnership communication). Sociocultural constructivist theories of communication describe social influences and communication as symbiotic. \cite{intro5, intro6, intro7} To date, the PV has rarely been factored into patient-centered outcome research (PCOR) methods. Absence of the PV limits the holistic understanding of patient care and can adversely influence decision making of patients, providers, and policymakers pursuing high value medical care. \cite{intro8,intro9, intro10} There is a critical need to extract, synthesize and integrate the PV buried in unstructured text to better inform PCOR.  

Despite this need, analyzing PV at scale remains challenging. The manual abstraction of the PV, represented by both facets PC and SDoH, from textual data and transforming it into a readily accessible structured format (for clinical and research use) is a laborious and expensive task \cite{intro11, intro12, intro13, intro14, intro15, intro16, intro17, roter1997rias}. Machine learning methods have been utilized to fill this gap. For example, with respect to PC, these methods have been applied to SM data to extract and code communication categories/domains like information seeking/sharing \cite{intro19, intro20, intro21, intro22, intro23, intro24, intro25, intro26, intro27, intro28, intro29}. These studies, however, failed to extract the more critical PC domains of the PV (e.g., shared-decision making and partnership). In addition, the coded PC categories/domains are limited in terms of specificity and scope. Each SM was coded to one mutually exclusive PC category despite the qualitative analyses \cite{intro17, intro30} establishing that a single SM could include contents of multiple PC categories. This singular coding may neglect the broader social context of SM and limits recognizing the intersecting clinical and social needs of patients. With respect to SDoH, existing machine learning studies mainly analyzed clinical notes in the electronic health records \cite{intro31, intro32, intro33, intro34, intro35, intro36,intro37}. These studies are limited not only because they ignored a rich platform of communication between patients and providers (i.e. SM) but also because they targeted one or two SDoH domains at the most in the clinical notes, which marks gaps in SM and machine learning research.

Recent advances in natural language processing (NLP) offer promising opportunities for modeling PV at scale. Topic modeling methods such as BERTopic \cite{grootendorst2022bertopic} can identify broader thematic structure, while few-shot and transformer-based methods \cite{intro39, intro40,wang2022dependency,wang2022conditional} enable fine-grained classification from limited annotated data. Yet, these techniques have typically been applied separately or trained on corpora misaligned with patient-facing language.

Together, prior work demonstrates the absence of a unified computational approach capable of modeling the full spectrum of PV. Existing models typically address either PC or SDoH in isolation, rely on single-label classification, or employ pretrained encoders that are not adapted to patient-authored narratives. These limitations prevent current methods from capturing the rich, overlapping, and contextually grounded expressions of medical concerns, social circumstances, and interpersonal communication that appear in SM. A domain-adapted, integrated approach that unifies contextual modeling, thematic representation, and multi-label prediction is therefore needed to support high-fidelity analysis of PV.

To address these gaps, we introduce PVminer, a domain-specific NLP framework for identifying, categorizing, and quantifying patient voice expressions in qualitative text data such as SM, interview and survey data. PVminer integrates large-scale domain-adapted pre-training, topic-informed representation learning, and a topic-augmented fine-tuning pipeline designed specifically for patient-generated text. This study builds on prior methodological work in extracting PC categories from SM \cite{intro42, intro43, raisa2023identifying} while extending beyond these efforts to jointly model communicative behaviors and SDoH expressions across multiple levels of granularity.

In this study, we make three primary contributions.
First, we introduce a comprehensive patient voice detection task that captures both communicative behaviors and SDoH-related expressions in a multi-label setting, reflecting the layered and overlapping nature of patient-authored communication.
Second, we present PVminer, an integrated NLP framework that combines domain-adapted pre-training, topic-informed representation learning, topic-augmented fine-tuning, and a unified evaluation protocol tailored to patient-generated text. An ablation study quantifies the contribution of each component.
Third, we develop a suite of domain-specific pre-trained models, including PV-BERT-base, PV-BERT-large, and PV-Topic-BERT, and demonstrate the benefits of patient-specific pre-training and topic augmentation for this task.
Fourth, we demonstrate the extensibility of PVminer by incorporating additional domain-related BERT variants beyond our own pre-trained models. Comparison experiments highlight the generalization of the PVminer framework and the performance advantages of our domain-specific encoders. 
Together, these contributions establish a methodological foundation for computational modeling of the patient voice. The overall workflow is shown in Fig. \ref{flowchart}.

\section{Related Work}\label{sec2}


Early research on patient-centered communication (PCC) relied heavily on manual qualitative frameworks such as the Roter Interaction Analysis System (RIAS) to analyze clinical encounters and secure messages \cite{intro17,roter1997rias,ref136}. These approaches provided rich conceptual categories,including information giving, partnership building, and emotional exchange, but required extensive human coding and lacked scalability. As electronic communication expanded, studies of patient–provider email and portal messages documented improved continuity and engagement \cite{intro30,ref86,ref87,ref88,ref89}. However, most analyses were descriptive or manually coded, offering limited insight into the linguistic and psychosocial structures embedded in patient-authored messages. The increasing volume of patient-generated data has thus outpaced the feasibility of traditional qualitative approaches.

\subsection{Machine Learning for Patient Message Classification}





 Early computational systems employed rule-based algorithms and traditional supervised classifiers, such as logistic regression and support vector machines, to categorize message intents, clinical queries, or emotional expressions \cite{intro13, intro14, intro15, intro16}. These studies demonstrated the feasibility of approximating human-coded annotations and revealed major functional categories such as medication management, information seeking, and symptom reporting.

The transition to deep learning brought convolutional and recurrent neural architectures capable of modeling sequential dependencies and contextual nuance \cite{intro29, intro39, intro40}. These models improved accuracy but remained constrained by limited annotated datasets and narrow generalizability across healthcare domains. Topic modeling frameworks, including latent Dirichlet allocation (LDA) and related extensions \cite{intro19, intro20, intro21, intro22, intro23, intro24, intro25}, enabled unsupervised discovery of thematic structures within patient messages. However, they typically assumed a single dominant topic per message and therefore could not capture the co-occurrence of multiple communicative intents that frequently appear in patient-authored narratives.

Despite these methodological advances, most approaches continued to treat patient communications as single-label units or narrowly scoped tasks, neglecting the layered nature of patient voice.

\subsection{Extracting Social Determinants of Health (SDOH) from Clinical Text}

Progress has occurred in the extraction of social determinants of health (SDoH) from unstructured electronic health record (EHR) data. Early work relied on lexicon- or ontology-driven methods to identify socioeconomic risks and contextual factors \cite{intro31,intro32,intro33}. With the advent of transformer architectures, models such as BERT and BioBERT enabled contextualized representations that substantially improved performance in biomedical NLP tasks \cite{devlin2019bert,ref83}. The emergence of standardized benchmarks, such as the 2022 n2c2/UW shared task on SDoH extraction \cite{ref81}, further advanced the field by providing curated datasets and transformer-based baselines, enabling progress in entity recognition, relation extraction, and risk profiling \cite{ref84,ref85}. However, these developments have focused almost exclusively on provider-authored notes. As a result, existing SDoH extraction frameworks largely overlook patient-provider communications, which contain self-reported lived experiences, social stressors, and environmental challenges that are often absent from clinical documentation.

\subsection{Gaps in Joint Modeling of Communication and Social Context}

Emerging interdisciplinary studies highlight the intertwined nature of communicative behaviors and social context in patient narratives. Research from the ECLIPPSE project \cite{ref90} and work by Yin et al. \cite{ref91} showed that linguistic features from secure messages can reveal patient literacy levels, emotional distress, or treatment adherence. Topic modeling of clinical notes has been used to uncover disease trajectories and social influences \cite{intro37}, while BERTopic \cite{grootendorst2022bertopic} introduced class-based TF–IDF representations that enhance interpretability of thematic structures. Recent zero-shot and few-shot learning studies \cite{intro39,intro40,ref92,ref93} demonstrate the feasibility of adapting pre-trained models to new health communication tasks with minimal annotation effort.

Despite these promising advances, existing computational research typically isolates communicative behaviors or SDoH rather than modeling them jointly. This separation stands in contrast to sociocultural constructivist theories and the Structural Influence Model (SIM), which emphasize the deep interdependence of communication, social structure, and health outcomes \cite{intro5,intro6,intro7,ref94,ref97}. Few NLP frameworks integrate these dimensions within a unified modeling paradigm, and even fewer address the overlapping, multi-label nature of patient-authored text.

Together, these limitations underscore the need for domain-adapted frameworks that integrate contextual language modeling, thematic structure, and multi-label prediction to capture the full breadth of patient voice. PVminer is designed to fill this methodological gap.

\section{Materials and Methods }


\begin{table*}[!t]
\centering
\caption{Demographic Distribution across Datasets. Some short terms are used: For the Sex, M is Male and F is Famale; For the Race, W is White, B is Black, A is Asian and O is Others.}
\label{tab:demographics}
\renewcommand{\arraystretch}{1.1}
\begin{tabular}{l|cc|cccc|c}
\hline
\textbf{Dataset} & \textbf{Sex (M)} & \textbf{Sex (F)} & \textbf{Race(W)} & \textbf{Race(B)} & \textbf{Race(A)} & \textbf{Race(O)} & \textbf{Ethnicity(Hispanic)} \\ 
\hline
\textbf{YNHH} & 81 (38\%) & 132 (62\%) & 152 (71\%) & 35 (16\%) & 11 (5\%) & 15 (7\%) & 30 (14\%) \\
\textbf{TXACC Woven} & 29 (48\%) & 32 (52\%) & 8 (13\%) & 6 (10\%) & 4 (7\%) & 43 (70\%) & 24 (39\%) \\
\textbf{TXACC Bethesda} & 71 (47\%) & 79 (53\%) & 20 (13\%) & 15 (10\%) & 9 (6\%) & 106 (71\%) & 60 (40\%) \\
\textbf{Survey} & 38 (26\%) & 109 (74\%) & 109 (74\%) & 8 (5\%) & 19 (13\%) & 11 (8\%) & 8 (5\%) \\
\textbf{Total} & 219 (38\%) & 352 (62\%) & 289 (51\%) & 64 (11\%) & 43 (8\%) & 175 (31\%) & 122 (21\%) \\
\hline
\end{tabular}
\end{table*}

\begin{table}[!t]
\centering
\caption{The statistics of the annotated secure message}
\label{tab:message_stats}
\small
\renewcommand{\arraystretch}{1.1}
\begin{tabular}{l|l}
\hline
\textbf{Item} & \textbf{Information} \\
\hline
\#Messages & 1137 \\
\#Provider's Message & 380 \\
\#Patient's Message & 757 \\
\#Total Words & 46038 \\
Ave. Words $\pm$ std & 40.49 $\pm$ 32.84 \\
Max Words & 261 \\
\hline
\end{tabular}
\end{table}

This section outlines the core NLP task for PVminer. We first introduce the multi-institutional datasets and its demographic composition, followed by the annotation design. Second, we show the hierarchical schema used to characterize patient voice elements. The detailed definitions are shown in Appendix A. Third, we formally define the multi-label multi-class classification task that underlies PVminer’s learning objective. Then, we show the methodology.

\subsection{Dataset Overview and Demographics}

Our analysis is based on a corpus of patient-generated communication text compiled from multiple data sources, including electronic patient messages and survey responses collected through ongoing patient-centered outcomes research initiatives. Some provider responses in the electronic patient-provider communication system are also included. Each document represents a unique instance of the patient voice, capturing individuals’ patient-provider communication dynamics.


Our annotated dataset comprises patient-generated messages collected from multiple sources, including secure patient-provider communications from Yale New Haven Health (YNHH), charitable clinics affiliated with the Texas Association for Charitable Clinics (TXACC), and patient survey responses. The TXACC data include messages from Woven Health and Bethesda Health Clinic, two charitable clinics based in Texas that serve diverse patient populations and contribute linguistic and cultural variability to the dataset. Across all sources, the annotated corpus includes 571 unique patients and a smaller number of providers, the majority of whom are affiliated with YNHH. 

Table~\ref{tab:demographics} summarizes the demographic composition of the annotated dataset. The corpus includes 571 unique patients representing diverse racial, gender, and ethnic backgrounds, with 62\% identifying as female and 21\% reporting Hispanic ethnicity. Racial distribution included 51\% White, 11\% Black or African American, 8\% Asian, and 30\% Other or mixed race. This diversity ensures that PVminer models capture a broad range of linguistic and cultural expressions of health experience. 

Table~\ref{tab:message_stats} presents statistics for the annotated subset used in the model development. The dataset comprises 1,137 patient-authored text totaling 46,038 words, with an average of 40.5~$\pm$~32.8 words per message and a maximum length of 261 words. Because each participant may contribute multiple single messages or threads of messages, message counts exceed the number of unique patients. This dataset provides sufficient linguistic variability to support robust fine-tuning and generalization of domain-specific transformer models.

\begin{figure*}[!t]
\centering
\includegraphics[width=0.85\textwidth]{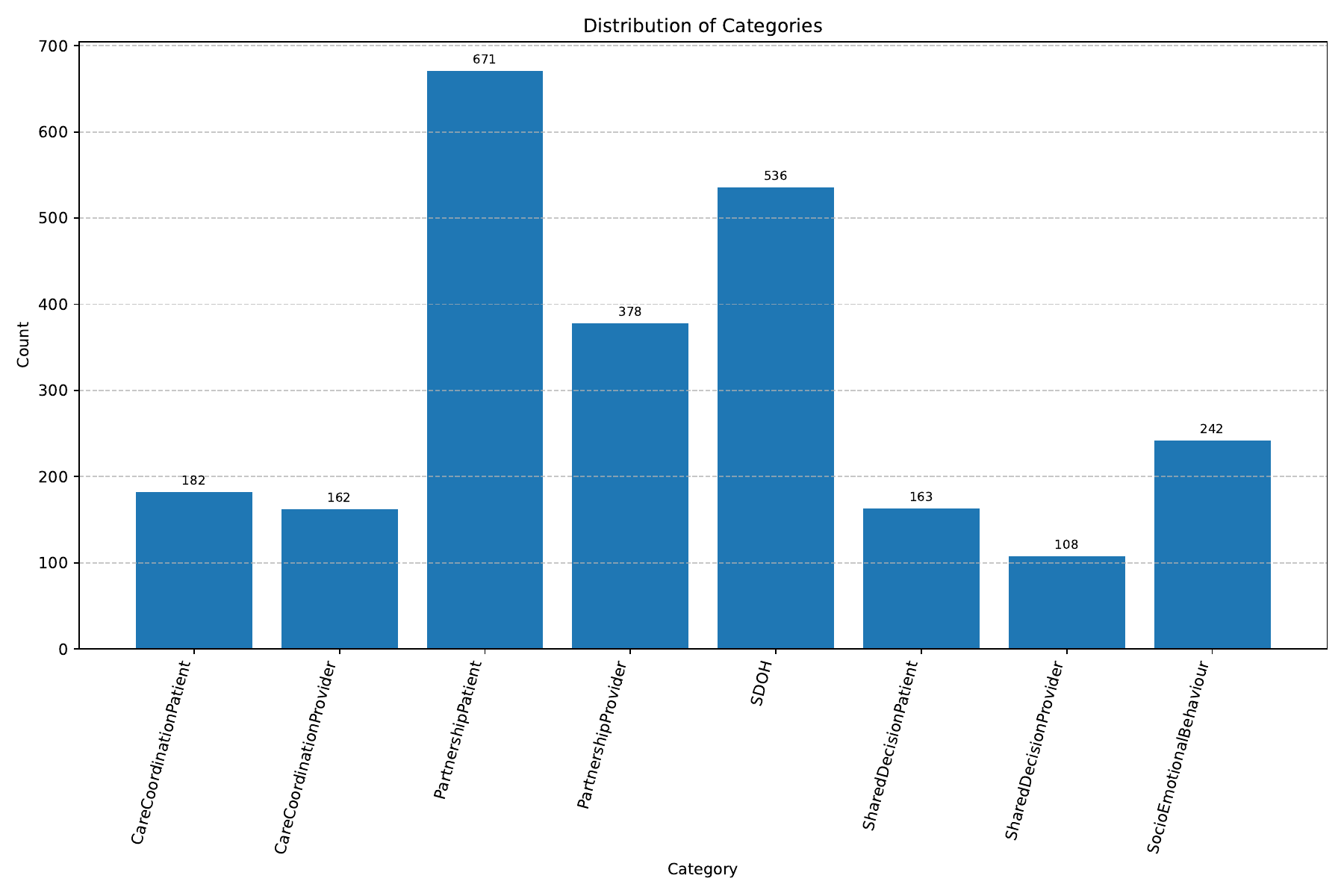}
\caption{Distribution of Code categories across all annotated patient messages. Each message may contain multiple non-overlapping labels.}\label{fig:code_dist}
\end{figure*}

\begin{figure*}[!t]
\centering
\includegraphics[width=0.85\textwidth]{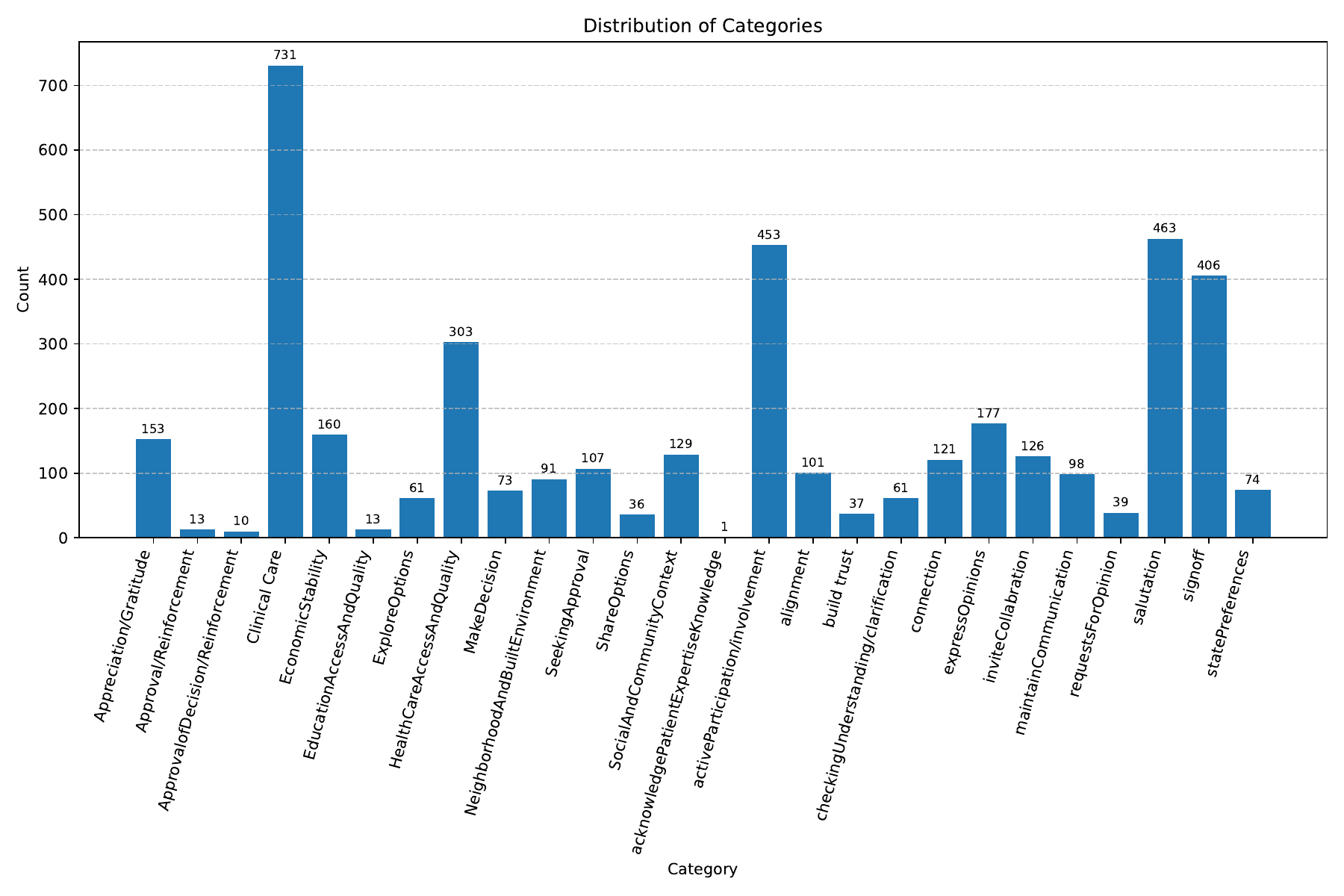}
\caption{Distribution of Subcodes representing fine-grained PV categories.}\label{fig:subcode_dist}
\end{figure*}

\begin{figure*}[!t]
\centering
\includegraphics[width=0.85\textwidth]{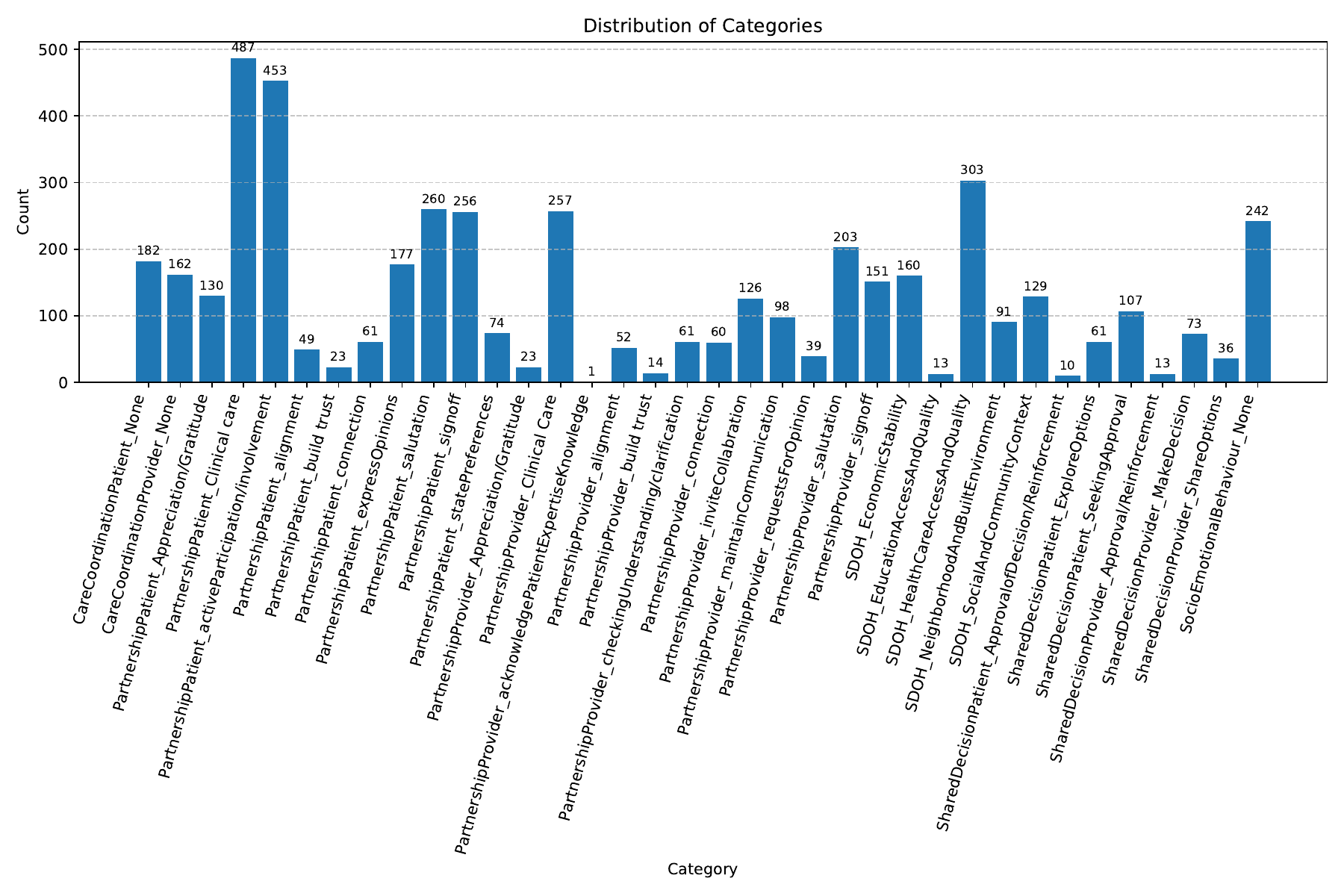}
\caption{Distribution of Combo labels (Code–Subcode pairs) in the annotated dataset. Absent Subcodes are labeled as None.}\label{fig:combo_dist}
\end{figure*}

\subsection{PV Codebook and Annotation}\label{subsec:annotation}


We developed a new PV-coding manual and lexicon that we used to annotate the messages and the survey data. Each message was annotated by domain experts trained in health communication and informatics using the eHOST platform \cite{yale2025annotation}. To ensure consistency, annotation proceeded iteratively through calibration meetings and consensus reviews. Difficult or ambiguous cases were resolved through group discussion, and the definitions of codes were refined through weekly quality-control sessions.

The final PV coding manual schema captures the multidimensional nature of the patient voice through a two-level hierarchy:

\begin{itemize}
    \item \textbf{Codes:} High-level domains representing broad communicative or social functions of the PV (e.g., Partnership, Shared Decision-Making, Socioemotional Support, and Social Determinants of Health).
    \item \textbf{Subcodes:} Fine-grained categories specifying the intent or contextual nuance within each Code (e.g., Appreciation, Connection, Economic Stability, or Access to Care).
\end{itemize}

Each unique combination of a Code and Subcode forms a \textbf{Combo} label. The PV manual including codes and subcodes along with there definitions are described in details in Appendix A. Figure~\ref{fig:code_dist}–\ref{fig:combo_dist} summarize the empirical distributions of Codes, Subcodes, and Combos in the annotated corpus. 

\subsection{Task Formulation}




We formulate patient voice extraction as a multi-label, multi-class classification task, reflecting the fact that a single patient message may simultaneously convey multiple communicative behaviors and social determinants. 

Let $M = \{m_1, m_2, \ldots, m_n\}$ denote the corpus of patient-authored messages, and let $\mathcal{L} = \{l_1, l_2, \ldots, l_L\}$ represent the complete set of \textit{Combo} categories, where each Combo corresponds to a unique pairing of a Code and a Subcode. For each message $m_i$, the model produces a vector of real-valued logits
\[
z_i = [z_{i,1}, z_{i,2}, \ldots, z_{i,L}],
\]
where each element $z_{i,j}$ reflects the model’s confidence for assigning label $l_j$ to message $m_i$.

An element-wise sigmoid function is applied to transform logits into probabilities,
\[
p_i = [p_{i,1}, p_{i,2}, \ldots, p_{i,L}],
\]
where each $p_{i,j} \in [0,1]$ represents the estimated likelihood that message $m_i$ expresses the communicative or social intent associated with Combo $l_j$.

Unlike single-label classification, this formulation allows multiple labels to be active for the same message, capturing the co-occurrence of related behaviors. For example, a single message may simultaneously convey appreciation (Partnership: Gratitude) and describe a financial challenge (SDOH: Economic Stability). To obtain discrete predictions, a fixed threshold (0.5) is applied to each probability, yielding a multi-hot label vector
\[
\hat{y}_i \in \{0,1\}^L
\]

The overall learning objective is therefore to estimate a mapping
\[
f: M \rightarrow \{0,1\}^L
\]
which assigns each message a multi-hot label vector over all possible PV categories. Predicted Combo labels are subsequently decomposed into their corresponding Code and Subcode representations for downstream analysis and evaluation.

This formulation enables systematic, large-scale quantification of patient voice by converting nuanced, context-dependent narratives into structured representations that can be compared and analyzed across individuals and populations.

\begin{table*}[!t]
\centering
\caption{Representative topics learned by the PV-Topic-BERT model. Topics reflect clinical, communication, and social themes commonly found in patient-authored messages.}
\label{tab:topics}
\small
\renewcommand{\arraystretch}{1.15}
\begin{tabular}{p{1.4cm} p{13.3cm}}
\hline
\textbf{Topic ID} & \textbf{Top Keywords} \\
\hline
0   & bloodwork, draw, blood, quest, orders, lab, station, tests \\
1   & vaccine, booster, flu, shot, vaccination, pfizer, moderna \\
3   & stool, diarrhea, bowel, constipation, abdominal, movements \\
4   & thyroid, tsh, levothyroxine, synthroid, mcg, thyroidectomy \\
5   & uti, bladder, urine, culture, urinary, antibiotics, burning \\
10  & mri, radiology, imaging, brain, scan, appointment \\
14  & welcome, appreciate, thank, perfect, much, great \\
20  & works, 3pm, 11am, thursday, noon, anytime, schedule \\
26  & foods, vegetables, oil, cheese, meals, diet, nausea \\
29  & authorization, insurance, approved, covered, denied, company \\
\hline
\end{tabular}
\end{table*}


\subsection{PVminer Framework Overview}

To provide a clear overview of our methodology, we first describe the pre-training of PV-BERT models, which are domain-adapted language models trained on a large corpus of unlabeled patient-authored messages. We then introduce PV-Topic-BERT, a topic-informed encoder constructed using BERTopic \cite{grootendorst2022bertopic} representations derived from the same corpus to capture latent thematic structure. Next, we describe the topic-augmented fine-tuning pipeline used to adapt all pre-trained models, including baseline BERT variants, for multi-label patient voice classification. These components collectively form PVminer. The overall workflow is illustrated in Fig.~\ref{flowchart}. While PVminer is optimized using our pre-trained PV-BERT models, its modular architecture allows any BERT-family encoder to be substituted into the same pipeline.

\subsection{Pre-training Domain-Adapted BERT Models: PV-BERT-base and PV-BERT-large}

BERT, a general-purpose language model, was originally trained on English Wikipedia and BookCorpus. Building on this foundation, numerous domain-adapted variants have been introduced, trained on biomedical, clinical, or scientific corpora. However, these corpora contain domain-specific language that differs substantially from the conversational and context-rich nature of patient-authored communication. Furthermore, the conceptual complexity of patient-generated messages and the limited availability of labeled data motivate the need for leveraging large-scale unlabeled in-domain text for effective domain adaptation.

To address this challenge, we pre-trained two domain-specific language models using a corpus of 
unlabeled patient-provider messages collected from YNHH. These messages varied widely in length, with the longest exceeding 2,750 words. For preprocessing, we applied the Punkt sentence tokenizer \cite{kiss2006sentence} from the Natural Language Toolkit (NLTK) \cite{bird2009nltk} to segment messages into shorter text units. This procedure produced 6,764,518 segments, each treated as an independent training instance. To satisfy BERT’s maximum input length of 512 tokens (including special tokens), we filtered all segments to a maximum of 510 tokens. A small number of very long segments could not be split further and were removed. 

We initialized pre-training from two standard checkpoints, bert-base-uncased and bert-large-uncased \cite{alsentzer2019clinicalbert,devlin2019bert}, producing two domain-adapted variants: PV-BERT-base and PV-BERT-large. Tokenization was performed using the WordPiece algorithm \cite{ouyang2022training}, which mitigates out-of-vocabulary issues by decomposing rare words into subword units. Both models were trained using the masked language modeling objective for 600 K iterations with an initial learning rate of $5 \times 10^{-5}$, weight decay of 0.01, and a masking probability of 0.15. An iteration corresponds to one optimizer update step.

Training was conducted on two NVIDIA A100 GPUs (80 GB each), with a batch size of 64 per device. Under these settings, PV-BERT-base converged in 3 days, while PV-BERT-large required 7 days. Unless otherwise noted, all remaining hyperparameters followed standard defaults.

\subsection{Pre-training Topic Model: PV-Topic-BERT}

To capture latent thematic structure in patient-provider messages, we constructed a domain-specific topic model using the BERTopic framework \cite{ouyang2022training}. BERTopic integrates transformer-based embeddings, dimensionality reduction through UMAP, density-based clustering with HDBSCAN, and class-based TF-IDF to generate coherent and interpretable topics from unstructured text. For sentence-level embeddings, we employed the pre-trained MiniLM-L6-v2 model from the Sentence Transformers library, which provides an effective balance between representational quality and computational efficiency.

The model was trained on 500 K unlabeled and non-deidentified patient-generated messages collected from the same health system as our annotated dataset. This corpus reflects authentic, real-world patient communication and provides a rich foundation for unsupervised topic discovery.

Unlike traditional topic models such as LDA, which require specifying the number of topics in advance, BERTopic leverages HDBSCAN to automatically infer the number of clusters from the density of the embedded space. Prior to clustering, UMAP was applied to reduce dimensionality while preserving local semantic structure, facilitating the ability of HDBSCAN to identify coherent clusters. The resulting model excluded outlier topics labeled as -1. Through pre-training, we get the PV-Topic-BERT model.

This pre-trained PV-Topic-BERT model serves two roles within PVminer. First, it is used to obtain topic distributions and class-based TF-IDF representations that augment encoder representations during fine-tuning. Second, these topic assignments are used in the inference process, enabling the classifier to jointly understand richer context and broader thematic structure to make the decision. A sample of representative topics is shown in Table \ref{tab:topics}, illustrating clinically meaningful themes such as laboratory testing, imaging, medication management, appointment coordination, and COVID-related inquiries.

\subsection{Fine-tuning PV-BERT-base and PV-BERT-large for Patient Voice Classification}

Following pre-training, all PVminer components were fine-tuned on the downstream task of patient voice classification. This task was formulated as a multi-label multi-class sequence classification problem, where each message may express multiple communicative behaviors, social factors, or combined patterns. The annotated dataset described in Table \ref{tab:message_stats} was randomly split into training and testing sets using an 80/20 ratio. To make sure the occurrence of all the classes, we applied iterative stratification \cite{sechidis2011stratification}, an extension of stratified sampling for multi-label data.

For input preparation, messages were tokenized using the same WordPiece tokenizer as the pre-trained encoders. All inputs remained within the 510-token limit in the training; thus, no truncation was needed. This ensured that full message content was provided to the model during fine-tuning.

To construct the output space, all unique combinations of Codes and corresponding Subcodes were mapped to integer indices. Each message was encoded as a multi-hot vector indicating the presence of one or more labels. During inference, predicted binary vectors were converted back to their label representations and decomposed into their corresponding Code, Subcode and Combo components for evaluation.

Fine-tuning was implemented using the Hugging Face Transformers library. Models were optimized with a learning rate of $5 \times 10^{-5}$, a batch size of~8, weight decay of 0.01, and linear learning rate decay with a warm-up ratio of 0.1. Training ran for up to 75 epochs with early stopping based on validation F1 score. To maintain interpretability and reproducibility, we explicitly specified the label2id and id2label mappings in the model configuration.

Each message includes metadata indicating whether it was written by a patient or a provider. We encoded this information by prepending a special token (e.g., \texttt{[PATIENT]} or \texttt{[PROVIDER]}) to the message text before tokenization. This allows the model to capture role-specific communication patterns, such as patient needs versus provider directives, and improves classification in short or ambiguous messages. Empirically, including author identity yielded consistent performance gains (see the Ablation study section).

To integrate thematic information, we augmented each message with topic cues derived from our PV-Topic-BERT model. For every input message, PV-Topic-BERT assigned the most likely topic, from which we extracted the top three keywords. These keywords were appended to the message text before tokenization, providing high-level semantic signals that enrich sparse or context-limited inputs.

Topic keywords help disambiguate short messages by linking them to broader themes learned from unlabeled data. For example, a message such as ``I am not feeling well today'' may be ambiguous on its own, but becomes more informative when paired with topic keywords such as ``chemo, fatigue, nausea.'' By combining surface-level text with latent thematic cues, PVminer leverages both explicit linguistic content and underlying semantic structure to improve multi-label classification performance.

Overall, this fine-tuning pipeline enables PVminer to adapt domain-specific language representations to the multi-layered nature of patient voice expression, integrating contextual embeddings, author identity, and topic-derived semantic structure.

\section{Experiments and Results}

\subsection{Experimental Settings}

We compare PVminer with both components PV-Topic-BERT and variants PV-BERT-base and PV-BERT-large against a range of widely used pre-trained transformer models from general-domain, biomedical, clinical, and scientific NLP. Baseline encoders include BERT-base and BERT-large \cite{devlin2019bert,alsentzer2019clinicalbert}, BioBERT \cite{lee2020biobert}, ClinicalBERT \cite{alsentzer2019clinicalbert}, SapBERT \cite{liu2021sapbert}, SciBERT \cite{beltagy2019scibert}, and TwHIN-BERT \cite{zhang2022twhinbert}. All baseline models are obtained from their official Hugging Face, and the identifiers reported in Tables~\ref{tab:code}--\ref{tab:combo} (e.g., \texttt{dmis-lab/biobert-v1.1}) correspond exactly to the official model repositories. We attempted to fine-tune \texttt{Twitter/twhin-bert-large}; however, the model failed to converge across multiple trials and was therefore excluded. Only the base version of TwHIN-BERT is reported.

All models are fine-tuned for up to 75 epochs using the default training configuration unless otherwise specified. The results in Tables~\ref{tab:code}--\ref{tab:combo} thus represent PVminer instantiated with different pre-trained encoders.

This experimental setup enables two complementary evaluations: (1) assessing the effectiveness of domain-specific pre-training combined with topic-aware augmentation for structured patient voice detection, and (2) comparing PV-BERT-base and PV-BERT-large against general-domain, biomedical, and clinical pre-trained models to demonstrate both the benefits of in-domain adaptation and the extensibility of the PVminer framework across related NLP settings.

\subsection{Evaluation Metric}


We evaluate model performance using micro-averaged Precision, Recall, and F1-score, which are well suited for imbalanced multi-label classification settings where label frequencies vary substantially. These metrics aggregate contributions from all labels and all instances, providing a global assessment of classification performance.

Each metric is computed across all message-level predictions as follows:

\begin{equation}
    precision = \frac{\sum_i |\hat{y}_i \cap y_i|}{\sum_i |\hat{y}_i|}
\end{equation}

\begin{equation}
    recall = \frac{\sum_i |\hat{y}_i \cap y_i|}{\sum_i |y_i|}
\end{equation}

\begin{equation}
    F1 = \frac{2 \times precision_{\mu} \times recall_{\mu}}{precision_{\mu} + recall_{\mu}}
\end{equation}

Here, $\hat{y}_i$ and $y_i$ denote the predicted and gold label sets for message $m_i$, respectively. 
We report these metrics separately for each prediction task: \textit{Code classification}, \textit{Subcode classification}, and \textit{Combo classification}, all of which are formulated as multi-label prediction problems.








\subsection{Comparing PVminer Performance  vs Other Models}

\subsubsection{PVminer - Code Classification}

\begin{table*}[!t]
\centering
\caption{Model performance on Code-level classification. Each reported result is averaged over five independent runs. Mean and standard deviation are provided to indicate the variability of each method. The best mean value in each column is highlighted in bold. All models are inside the PVminer.}
\label{tab:code}
\small
\renewcommand{\arraystretch}{1.1}
\begin{tabular}{>{\raggedright\arraybackslash}p{3.8cm}|cc|cc|cc}
\hline
\textbf{Model} & \textbf{Precision$_{mean}$} & \textbf{Precision$_{std}$} &
\textbf{Recall$_{mean}$} & \textbf{Recall$_{std}$} & \textbf{F1$_{mean}$} & \textbf{F1$_{std}$} \\
\hline
\makecell[l]{dmis-lab/biobert-\\v1.1} & 86.95 & 0.24 & 71.78 & 0.70 & 78.64 & 0.49 \\
\hline
\makecell[l]{allenai/scibert\_\\scivocab\_uncased} & 85.96 & 0.45 & 72.50 & 0.95 & 78.65 & 0.38 \\
\hline
\makecell[l]{emilyalsentzer/\\Bio\_ClinicalBERT} & 86.52 & 0.44 & 70.64 & 0.86 & 77.77 & 0.39 \\
\hline
\makecell[l]{cambridgeltl/SapBERT-\\from-PubMedBERT-fulltext} & 84.93 & 0.30 & 71.90 & 0.52 & 77.87 & 0.42 \\
\hline
bert-base-uncased & 88.52 & 0.64 & 74.61 & 0.47 & 80.97 & 0.05 \\
\hline
bert-large-uncased & 86.01 & 0.70 & 75.15 & 1.77 & 80.21 & 1.25 \\
\hline
\makecell[l]{Twitter/twhin-\\bert-base} & 85.56 & 1.21 & 74.37 & 0.39 & 79.57 & 0.47 \\
\hline
PV-BERT-base & 88.75 & 1.04 & 74.85 & 0.68 & \textbf{81.20} & 0.30 \\
\hline
PV-BERT-large & 89.47 & 0.62 & 76.11 & 0.95 & \textbf{82.25} & 0.33 \\
\hline
\end{tabular}
\end{table*}

Table~\ref{tab:code} reports model performance on the Code-level classification task within the unified PVminer framework. Because all encoders are evaluated using the same topic-augmented architecture, the results reflect differences between pre-training strategies while simultaneously illustrating the overall effectiveness of PVminer itself.

Across all encoders, PVminer achieves strong absolute performance, with every model reaching an F1-score around 80\%. 

Among the baseline encoders, the general-domain BERT models perform particularly well, with BERT-base achieving an F1-score of 80.97\% and BERT-large achieving 80.21\%. These values exceed those of biomedical and scientific transformers such as BioBERT (78.64\%), SciBERT (78.65\%), and SapBERT (77.87\%). This pattern suggests that conversational patient language aligns more closely with general-domain text than with biomedical literature, and it further highlights the importance of domain alignment in pre-training.

The strongest results come from our domain-specific encoders, PV-BERT-base and PV-BERT-large, which are pre-trained on large-scale patient-provider communication. The base variant achieves an F1-score of 81.20\%, surpassing all baseline models. The large variant delivers the best overall performance, reaching Precision = 89.47\%, Recall = 76.11\%, and F1 = 82.25\%. Because all models share the same PVminer architecture and topic augmentation, these improvements directly reflect the value of pre-training on patient-generated text.

Partnership behaviors were most frequent, reflecting patients’ strong emphasis on collaborative and informational exchanges. SDoH-related content, particularly Economic Stability and Access to Services, also appeared frequently, highlighting the contextual richness of patient narratives.

\subsubsection{PVminer - Subcode Classification}

\begin{table*}[!t]
\centering
\caption{Model performance on Subcode-level classification. Each reported result is averaged over five independent runs. Mean and standard deviation are provided to indicate the variability of each method. The best mean value in each column is highlighted in bold. All models are inside the PVminer.}
\label{tab:subcode}
\small
\renewcommand{\arraystretch}{1.1}
\begin{tabular}{>{\raggedright\arraybackslash}p{3.8cm}|cc|cc|cc}
\hline
\textbf{Model} & \textbf{Precision$_{mean}$} & \textbf{Precision$_{std}$} &
\textbf{Recall$_{mean}$} & \textbf{Recall$_{std}$} & \textbf{F1$_{mean}$} & \textbf{F1$_{std}$} \\
\hline
\makecell[l]{dmis-lab/biobert-\\v1.1} & 81.65 & 0.39 & 72.85 & 1.47 & 76.99 & 0.83 \\
\hline
\makecell[l]{allenai/scibert\_\\scivocab\_uncased} & 81.07 & 0.48 & 73.27 & 0.75 & 76.97 & 0.59 \\
\hline
\makecell[l]{emilyalsentzer/\\Bio\_ClinicalBERT} & 81.50 & 0.96 & 71.63 & 1.05 & 76.24 & 0.84 \\
\hline
\makecell[l]{cambridgeltl/SapBERT-\\from-PubMedBERT-fulltext} & 80.09 & 0.30 & 72.20 & 0.42 & 75.94 & 0.37 \\
\hline
bert-base-uncased & 81.90 & 0.93 & 74.80 & 1.33 & 78.18 & 0.95 \\
\hline
bert-large-uncased & 80.88 & 1.48 & 75.64 & 3.63 & 78.15 & 2.63 \\
\hline
\makecell[l]{Twitter/twhin-\\bert-base} & 81.14 & 0.79 & 76.52 & 0.09 & 78.76 & 0.33 \\
\hline
PV-BERT-base & 84.17 & 0.90 & 76.48 & 0.38 & \textbf{80.14} & 0.35 \\
\hline
PV-BERT-large & 82.58 & 0.53 & 77.28 & 0.38 & \textbf{79.84} & 0.19 \\
\hline
\end{tabular}
\end{table*}

Table~\ref{tab:subcode} reports the performance of all evaluated encoders on the Subcode-level classification task within the PVminer framework. As expected, performance drops slightly compared to Code-level prediction due to the finer granularity and larger label space of subcodes. Many subcodes capture subtle distinctions in communicative intent (e.g., different types of requests, updates, or contextual qualifiers), making this task inherently more challenging. Despite this increased difficulty, PVminer maintains strong overall performance across all encoder families, again demonstrating the robustness of the framework.

Among the baseline encoders, the general-domain BERT models provide the strongest results, with BERT-base achieving an F1-score of 78.18\% and BERT-large achieving 78.15\%. Biomedical and scientific domain models show slightly lower performance: BioBERT reaches an F1-score of 76.99\%, SciBERT yields 76.97\%, and SapBERT achieves 75.94\%. Although these models incorporate extensive biomedical knowledge, their pre-training corpora do not reflect the conversational and context-rich nature of patient-authored communication, which likely limits their ability to capture fine-grained subcode distinctions.

Our domain-specific models provide the highest overall performance. The PV-BERT-base encoder achieves the best results across all metrics, with Precision = 84.17\%, Recall = 76.48\%, and F1 = 80.14\%. The PV-BERT-large variant performs closely behind (F1 = 79.84\%), showing strong Recall (77.28\%) but slightly lower Precision. This pattern suggests that the larger model may capture broader semantic context, while the base variant provides a tighter fit for fine-grained communicative cues.

\subsubsection{PVminer - Combo classification}

\begin{table*}[!t]
\centering
\caption{Model performance on Combo-level classification. Each reported result is averaged over five independent runs. Mean and standard deviation are provided to indicate the variability of each method. The best mean value in each column is highlighted in bold. All models are inside the PVminer.}
\label{tab:combo}
\small
\renewcommand{\arraystretch}{1.1}
\begin{tabular}{>{\raggedright\arraybackslash}p{3.8cm}|cc|cc|cc}
\hline
\textbf{Model} & \textbf{Precision$_{mean}$} & \textbf{Precision$_{std}$} &
\textbf{Recall$_{mean}$} & \textbf{Recall$_{std}$} & \textbf{F1$_{mean}$} & \textbf{F1$_{std}$} \\
\hline
\makecell[l]{dmis-lab/biobert-\\v1.1} & 79.67 & 0.46 & 70.32 & 1.17 & 74.70 & 0.68 \\
\hline
\makecell[l]{allenai/scibert\_\\scivocab\_uncased} & 79.16 & 0.32 & 70.55 & 0.69 & 74.61 & 0.50 \\
\hline
\makecell[l]{emilyalsentzer/\\Bio\_ClinicalBERT} & 79.45 & 0.64 & 69.63 & 0.70 & 74.21 & 0.49 \\
\hline
\makecell[l]{cambridgeltl/SapBERT-\\from-PubMedBERT-fulltext} & 78.01 & 0.07 & 69.76 & 0.00 & 73.66 & 0.03 \\
\hline
bert-base-uncased & 80.74 & 0.36 & 72.63 & 0.92 & 76.47 & 0.52 \\
\hline
bert-large-uncased & 78.63 & 1.25 & 72.56 & 3.82 & 75.45 & 2.64 \\
\hline
\makecell[l]{Twitter/twhin-\\bert-base} & 79.21 & 0.79 & 73.48 & 0.09 & 76.24 & 0.34 \\
\hline
PV-BERT-base & 82.06 & 0.95 & 73.58 & 0.26 & \textbf{77.58} & 0.30 \\
\hline
PV-BERT-large & 81.53 & 0.45 & 74.54 & 0.57 & \textbf{77.87} & 0.17 \\
\hline
\end{tabular}
\end{table*}

Table~\ref{tab:combo} summarizes model performance on the Combo-level classification task, which evaluates the ability to jointly predict both Code and Subcode labels. 

Among the baseline encoders within the PVminer framework, the general-domain BERT-base model achieves the strongest performance (F1 = 76.47\%), followed closely by TwHIN-BERT (F1 = 76.24\%). BERT-large performs slightly lower, likely due to its tendency toward overfitting or sensitivity to the relatively small annotated dataset. Biomedical and scientific encoders again lag behind, with BioBERT (F1 = 74.70\%) and SciBERT (F1 = 74.61\%) performing comparably and SapBERT showing the lowest baseline performance (F1 = 73.66\%). These results reinforce the observation that models pre-trained on biomedical literature do not fully capture the conversational structure and narrative style of patient-generated messages.

Our domain-specific pre-trained encoders provide the strongest performance at this hierarchical level. The PV-BERT-base model attains Precision = 82.06\%, Recall = 73.58\%, and an overall F1-score of 77.58\%. The larger model, PV-BERT-large, achieves the highest Recall (74.54\%) and the best overall F1-score (77.87\%), demonstrating improved ability to capture the combined semantic and functional structure of composite labels. 

\subsection{Analyzing PV-BERT-large Performance}

\subsubsection{Code-level analysis}

\begin{table*}[!t]
\centering
\caption{Class-level performance metrics of the PV-BERT-large model on the Code-level classification task. Each reported result is averaged over five independent runs. Mean and standard deviation are provided to indicate the variability across runs.}

\label{tab:classlevel}
\small
\renewcommand{\arraystretch}{1.1}
\begin{tabular}{l|cc|cc|cc}
\hline
\textbf{Class Label} & \textbf{Precision$_{mean}$} & \textbf{Precision$_{std}$} & 
\textbf{Recall$_{mean}$} & \textbf{Recall$_{std}$} & \textbf{F1$_{mean}$} & \textbf{F1$_{std}$} \\
\hline
All & 89.47 & 0.76 & 76.11 & 1.16 & 82.25 & 0.41 \\
\hline
CareCoordinationPatient & 86.44 & 7.40 & 62.39 & 1.48 & 72.33 & 1.75 \\
\hline
CareCoordinationProvider & 76.60 & 1.56 & 68.57 & 2.86 & 72.35 & 2.02 \\
\hline
PartnershipPatient & 93.16 & 0.71 & 88.81 & 1.21 & 90.93 & 0.30 \\
\hline
PartnershipProvider & 97.97 & 1.86 & 86.81 & 1.10 & 92.04 & 0.62 \\
\hline
SDOH & 89.63 & 2.22 & 77.60 & 0.47 & 83.17 & 0.68 \\
\hline
SharedDecisionPatient & 89.07 & 3.86 & 60.95 & 1.65 & 72.32 & 0.16 \\
\hline
SharedDecisionProvider & 57.79 & 7.23 & 38.46 & 10.18 & 45.75 & 8.06 \\
\hline
SocioEmotionalBehaviour & 86.78 & 2.53 & 65.61 & 3.30 & 74.67 & 2.17 \\
\hline
\end{tabular}
\end{table*}


Table~\ref{tab:classlevel} reports the class-level performance of the domain-adapted PV-BERT-large encoder within the PVminer framework for Code-level classification. The model achieves strong overall performance (F1 = 82.25 \%), with high precision (89.47 \%) and robust recall (76.11 \%). Class-specific scores, however, vary substantially, reflecting differences in label frequency, semantic ambiguity, and the linguistic consistency of each communication category.


The highest-performing categories include PartnershipPatient (F1 = 90.93 \%) and PartnershipProvider (F1 = 92.04 \%). These categories tend to contain explicit and formulaic communicative acts such as acknowledgements, expressions of thanks, and confirmations (e.g., ``thank you for your help,'' ``I appreciate the explanation,'' or ``that makes sense''). Because these behaviors use consistent phrasing and appear frequently in the corpus, the model identifies them reliably. The model also performs well on SDOH (F1 = 83.17 \%), indicating strong ability to detect social and contextual signals embedded in patient narratives. Many SDOH expressions are lexically grounded, such as references to financial hardship or access barriers (e.g., ``I cannot afford the medication'' or ``I do not have transportation to the clinic''), which supports consistent recognition.

Performance is more moderate in categories with broader linguistic variability. CareCoordination\-Patient and Care\-Coordination\-Provider, with F1-scores around 72 \%, contain diverse expressions related to scheduling, referrals, logistics, and navigation of the health system. SocioEmotional\-Behaviour (F1 = 74.67 \%) shows similar patterns, consistent with the subtle and context-dependent nature of affective language.

The most challenging categories involve shared decision-making. SharedDecision\-Patient (F1 = 72.32 \%) already shows reduced performance, and SharedDecision\-Provider exhibits the lowest category-level result (F1 = 45.75 \%). These categories are both infrequent and conceptually complex. They often require inference about negotiations, treatment discussions, or implicit decision processes rather than explicit lexical cues, which contributes to higher variance across runs.

\subsubsection{SubCode Level Analysis}

\begin{table*}[!t]
\centering
\caption{Detailed class-level performance of the PV-BERT-large model on Subcode categories. Each reported result is averaged over five independent runs. Mean and standard deviation are provided to indicate the variability across runs.}
\label{tab:subcode_detail}
\small
\renewcommand{\arraystretch}{1.1}
\begin{tabular}{l|cc|cc|cc}
\hline
\textbf{Class Label} & \textbf{Precision$_{mean}$} & \textbf{Precision$_{std}$} &
\textbf{Recall$_{mean}$} & \textbf{Recall$_{std}$} & \textbf{F1$_{mean}$} & \textbf{F1$_{std}$} \\
\hline
All & 82.58 & 0.65 & 77.28 & 0.46 & 79.84 & 0.23 \\
\hline
Appreciation/Gratitude & 95.51 & 0.30 & 68.82 & 4.93 & 79.93 & 3.41 \\
\hline
Approval/Reinforcement & 0.00 & 0.00 & 0.00 & 0.00 & 0.00 & 0.00 \\
\hline
ApprovalofDecision/Reinforcement & 0.00 & 0.00 & 0.00 & 0.00 & 0.00 & 0.00 \\
\hline
Clinical Care & 82.91 & 1.59 & 91.61 & 0.65 & 87.03 & 0.59 \\
\hline
EconomicStability & 91.88 & 1.94 & 77.78 & 3.92 & 84.23 & 3.09 \\
\hline
EducationAccessAndQuality & 72.22 & 25.46 & 33.33 & 14.43 & 43.49 & 12.28 \\
\hline
ExploreOptions & 72.42 & 10.45 & 41.67 & 0.00 & 52.73 & 2.78 \\
\hline
HealthCareAccessAndQuality & 73.46 & 1.36 & 60.42 & 6.31 & 66.16 & 3.81 \\
\hline
MakeDecision & 55.42 & 5.05 & 33.33 & 10.41 & 41.11 & 8.39 \\
\hline
NeighborhoodAndBuiltEnvironment & 74.81 & 7.14 & 31.82 & 4.55 & 44.62 & 5.66 \\
\hline
SeekingApproval & 81.75 & 7.40 & 61.11 & 2.41 & 69.91 & 4.24 \\
\hline
ShareOptions & 43.98 & 6.26 & 41.67 & 7.22 & 42.47 & 4.79 \\
\hline
SocialAndCommunityContext & 85.31 & 1.16 & 83.33 & 7.43 & 84.23 & 4.43 \\
\hline
acknowledgePatientExpertiseKnowledge & 0.00 & 0.00 & 0.00 & 0.00 & 0.00 & 0.00 \\
\hline
activeParticipation/involvement & 75.23 & 1.04 & 90.11 & 1.90 & 81.99 & 1.39 \\
\hline
alignment & 93.33 & 2.89 & 77.78 & 2.41 & 84.85 & 2.62 \\
\hline
build trust & 0.00 & 0.00 & 0.00 & 0.00 & 0.00 & 0.00 \\
\hline
checkingUnderstanding/clarification & 71.11 & 18.36 & 28.57 & 7.14 & 40.70 & 10.07 \\
\hline
connection & 85.17 & 2.02 & 80.77 & 0.00 & 82.90 & 0.95 \\
\hline
expressOpinions & 74.17 & 1.44 & 72.81 & 3.04 & 73.43 & 0.80 \\
\hline
inviteCollabration & 64.77 & 0.65 & 63.33 & 3.33 & 64.02 & 1.78 \\
\hline
maintainCommunication & 82.01 & 6.42 & 77.78 & 2.75 & 79.73 & 3.12 \\
\hline
requestsForOpinion & 61.11 & 9.62 & 39.39 & 13.89 & 47.45 & 12.36 \\
\hline
salutation & 100.00 & 0.00 & 100.00 & 0.00 & 100.00 & 0.00 \\
\hline
signoff & 90.37 & 0.92 & 99.22 & 0.68 & 94.58 & 0.29 \\
\hline
statePreferences & 66.29 & 14.03 & 44.44 & 13.88 & 52.95 & 14.16 \\
\hline
\end{tabular}
\end{table*}

Table~\ref{tab:subcode_detail} presents detailed class-level results of the PV-BERT-large model on the secure message Subcode-level classification task. The model achieves a micro-averaged F1-score of 79.84 \%, with strong overall precision (82.58 \%) and recall (77.28 \%). Substantial variability appears across individual subcodes, reflecting differences in label frequency, semantic clarity, and the complexity of communicative intent within each category.

The model performs particularly well on well-structured and frequently observed behaviors such as salutation (F1 = 100.00 \%), signoff (F1 = 94.58 \%), clinical care (F1 = 87.03 \%), alignment (F1 = 84.85 \%), and connection (F1 = 82.90 \%). These subcodes often include consistent lexical or syntactic patterns, such as closing phrases, care instructions, or supportive alignment statements, which facilitate reliable identification. Strong performance is also observed in activeParticipation (F1 = 81.99 \%) and maintainCommunication (F1 = 79.73 \%), both of which capture patient engagement and ongoing conversational coordination.

In contrast, several subcodes show low or zero F1-scores, usually due to extreme label sparsity or diffuse linguistic boundaries. Approval\-slash\-Reinforcement, build\-trust, and acknowledgePatientExpertiseKnowledge receive no correct predictions because they contain too few instances for the model to learn meaningful patterns. Other subcodes show reduced but non-zero performance, including MakeDecision (F1 = 41.11 \%), ShareOptions (F1 = 42.47 \%), and EducationAccessAndQuality (F1 = 43.49 \%). These categories often involve abstract reasoning, implicit negotiation, or domain knowledge about health literacy and shared decision-making, which makes them difficult to detect from message text alone.

\begin{table*}[!h]
\centering
\caption{Detailed class-level performance of the PV-BERT-large model on Combo categories. Each reported result is averaged over five independent runs. Mean and standard deviation are provided to indicate the variability across runs.}
\label{tab:combo_detail}
\small
\renewcommand{\arraystretch}{1.1}
\begin{tabular}{>{\raggedright\arraybackslash}p{4.8cm}|cc|cc|cc}
\hline
\textbf{Class Label} & \textbf{Precision$_{mean}$} & \textbf{Precision$_{std}$} &
\textbf{Recall$_{mean}$} & \textbf{Recall$_{std}$} & \textbf{F1$_{mean}$} & \textbf{F1$_{std}$} \\
\hline
All & 81.53 & 0.56 & 74.54 & 0.70 & 77.87 & 0.21 \\
\hline
CareCoordinationPatient\_None & 86.44 & 7.40 & 62.39 & 1.48 & 72.33 & 1.75 \\
\hline
CareCoordinationProvider\_None & 76.60 & 1.56 & 68.57 & 2.86 & 72.35 & 2.02 \\
\hline
\makecell[l]{PartnershipPatient\_Appreciation/\\Gratitude} & 90.16 & 0.27 & 70.51 & 2.22 & 79.12 & 1.50 \\
\hline
\makecell[l]{PartnershipPatient\_Clinical\\care} & 86.16 & 2.14 & 88.12 & 0.99 & 87.12 & 1.39 \\
\hline
\makecell[l]{PartnershipPatient\_\\activeParticipation/involvement} & 75.23 & 1.04 & 90.11 & 1.90 & 81.99 & 1.39 \\
\hline
PartnershipPatient\_alignment & 96.30 & 6.42 & 83.33 & 14.43 & 89.16 & 11.24 \\
\hline
PartnershipPatient\_build trust & 0.00 & 0.00 & 0.00 & 0.00 & 0.00 & 0.00 \\
\hline
PartnershipPatient\_connection & 94.44 & 4.81 & 78.57 & 0.00 & 85.74 & 1.95 \\
\hline
PartnershipPatient\_expressOpinions & 74.17 & 1.44 & 72.81 & 3.04 & 73.43 & 0.80 \\
\hline
PartnershipPatient\_salutation & 98.06 & 1.92 & 96.15 & 0.00 & 97.09 & 0.94 \\
\hline
PartnershipPatient\_signoff & 83.07 & 0.71 & 99.35 & 1.13 & 90.48 & 0.52 \\
\hline
PartnershipPatient\_statePreferences & 66.29 & 14.03 & 44.44 & 13.88 & 52.95 & 14.16 \\
\hline
\makecell[l]{PartnershipProvider\_Appreciation/\\Gratitude} & 100.00 & 0.00 & 40.00 & 20.00 & 55.16 & 20.90 \\
\hline
\makecell[l]{PartnershipProvider\_Clinical\\Care} & 74.56 & 2.24 & 90.48 & 2.06 & 81.73 & 1.50 \\
\hline
\makecell[l]{PartnershipProvider\_acknowledge\\PatientExpertiseKnowledge} & 0.00 & 0.00 & 0.00 & 0.00 & 0.00 & 0.00 \\
\hline
PartnershipProvider\_alignment & 78.11 & 9.33 & 66.67 & 0.00 & 71.77 & 3.83 \\
\hline
PartnershipProvider\_build trust & 0.00 & 0.00 & 0.00 & 0.00 & 0.00 & 0.00 \\
\hline
\makecell[l]{PartnershipProvider\_\\checkingUnderstanding/clarification} & 71.11 & 18.36 & 28.57 & 7.14 & 40.70 & 10.07 \\
\hline
PartnershipProvider\_connection & 76.92 & 0.00 & 83.33 & 0.00 & 80.00 & 0.00 \\
\hline
\makecell[l]{PartnershipProvider\_invite\\Collabration} & 64.77 & 0.65 & 63.33 & 3.33 & 64.02 & 1.78 \\
\hline
\makecell[l]{PartnershipProvider\_maintain\\Communication} & 82.01 & 6.42 & 77.78 & 2.75 & 79.73 & 3.12 \\
\hline
\makecell[l]{PartnershipProvider\_requestsFor\\Opinion} & 61.11 & 9.62 & 39.39 & 13.89 & 47.45 & 12.36 \\
\hline
PartnershipProvider\_salutation & 95.68 & 0.05 & 98.52 & 1.28 & 97.08 & 0.65 \\
\hline
PartnershipProvider\_signoff & 93.84 & 2.98 & 89.22 & 1.70 & 91.47 & 2.25 \\
\hline
SDOH\_EconomicStability & 91.88 & 1.94 & 77.78 & 3.92 & 84.23 & 3.09 \\
\hline
\makecell[l]{SDOH\_EducationAccessAnd\\Quality} & 72.22 & 25.46 & 33.33 & 14.43 & 43.49 & 12.28 \\
\hline
\makecell[l]{SDOH\_HealthCareAccessAnd\\Quality} & 73.46 & 1.36 & 60.42 & 6.31 & 66.16 & 3.81 \\
\hline
\makecell[l]{SDOH\_NeighborhoodAndBuilt\\Environment} & 74.81 & 7.14 & 31.82 & 4.55 & 44.62 & 5.66 \\
\hline
SDOH\_SocialAndCommunityContext & 85.31 & 1.16 & 83.33 & 7.43 & 84.23 & 4.43 \\
\hline
\makecell[l]{SharedDecisionPatient\_\\ApprovalofDecision/Reinforcement} & 0.00 & 0.00 & 0.00 & 0.00 & 0.00 & 0.00 \\
\hline
\makecell[l]{SharedDecisionPatient\_Explore\\Options} & 72.42 & 10.45 & 41.67 & 0.00 & 52.73 & 2.78 \\
\hline
\makecell[l]{SharedDecisionPatient\_Seeking\\Approval} & 81.75 & 7.40 & 61.11 & 2.41 & 69.91 & 4.24 \\
\hline
\makecell[l]{SharedDecisionProvider\_Approval/\\Reinforcement} & 0.00 & 0.00 & 0.00 & 0.00 & 0.00 & 0.00 \\
\hline
\makecell[l]{SharedDecisionProvider\_Make\\Decision} & 55.42 & 5.05 & 33.33 & 10.41 & 41.11 & 8.39 \\
\hline
\makecell[l]{SharedDecisionProvider\_Share\\Options} & 43.98 & 6.26 & 41.67 & 7.22 & 42.47 & 4.79 \\
\hline
\makecell[l]{SocioEmotionalBehaviour\_\\None} & 86.78 & 2.53 & 65.61 & 3.30 & 74.67 & 2.17 \\
\hline
\end{tabular}
\end{table*}

\subsubsection{Combo-Level Analysis}

Table~\ref{tab:combo_detail} presents class-level results for the PV-BERT-large encoder within the PVminer framework on the Combo-level classification task. This task is the most challenging of the three evaluated settings because each label combines both a Code and a Subcode, resulting in a larger and more semantically complex label space. The model achieves an overall F1-score of 77.87 \%, with precision of 81.53 \% and recall of 74.54 \%, indicating strong performance despite the expanded output structure.

Performance varies across individual Combo categories. Behaviors with consistent lexical cues or frequent representation in the dataset show strong results. Examples include PartnershipPatient\_salutation (F1 = 97.09 \%), Partnership\-Patient\_signoff (F1 = 90.48 \%), PartnershipPatient\_clinical care (F1 = 87.12 \%), PartnershipPatient\_alignment (F1 = 89.16 \%), and PartnershipProvider\_signoff (F1 = 91.47 \%). These combinations tend to be expressed using stable phrasing patterns, which supports more reliable model detection.

The model also performs well on several SDOH-related combinations, such as SDOH\_EconomicStability (F1 = 84.23 \%) and SDOH\_SocialAndCommunityContext (F1 = 84.23 \%). These results suggest that the model effectively captures references to financial strain, community support, and related contextual indicators when embedded within broader communicative behaviors.

Performance decreases for Combo categories that are infrequent, semantically abstract, or linguistically diffuse. Several categories, such as PartnershipPatient\_build trust and SharedDecisionProvider\_Approval/Reinforcement, receive zero-valued metrics due to extremely small sample sizes. Other challenging combinations include Shared\-Decision\-Provider\_MakeDecision (F1 = 41.11 \%), Shared\-Decision\-
Patient\_ExploreOptions (F1 = 52.73 \%), and SDOH\_\-Neighborhood\-AndBuiltEnvironment (F1 = 44.62 \%). These combinations often require interpreting implicit reasoning or contextual inferences that may not be directly encoded in surface wording.

\subsection{Ablation Study of PVminer}

To assess the contribution of each component in PVminer, we conduct an ablation study using our pre-trained PV-BERT-large encoder as the base model. Three configurations are evaluated: (1) the encoder alone, (2) the encoder with author identity prepended as a special token (e.g., PATIENT or PROVIDER), and (3) the full PVminer configuration, which incorporates both author identity and topic keywords derived from the BERTopic model. This design isolates the effect of conversational metadata and topic-aware augmentation on the three classification tasks. Table~\ref{tab:ablation} summarizes the results.

The base PV-BERT-large encoder achieves F1-scores of 78.62 \% on Code-level prediction, 76.21 \% on Subcode-level prediction, and 73.94 \% on Combo-level prediction. Adding author identity yields consistent gains across all tasks, improving Code-level F1 to 79.21 \%, Subcode-level F1 to 77.01 \%, and Combo-level F1 to 75.33 \%. These improvements confirm that speaker role provides valuable contextual information, helping the model differentiate between patient-initiated requests and provider-delivered instructions.

The full PVminer configuration, which incorporates both author identity and topic keywords, achieves the highest performance on every task: 82.25 \% on Code-level prediction, 79.84 \% on Subcode-level prediction, and 77.87 \% on Combo-level prediction. Topic augmentation provides particularly substantial gains, indicating that high-level semantic cues derived from unlabeled messages complement encoder representations and enhance the model's ability to identify complex and overlapping communicative behaviors.

Overall, the ablation results show that both components contribute meaningfully to performance, and that the combination of domain-specific pre-training, author identity, and topic-aware augmentation is essential for achieving the highest accuracy in structured patient voice detection.

\begin{table*}[!t]
\centering
\caption{Ablation study of PVminer using the PV-BERT-large encoder. Results show mean and standard deviation across five runs with different random seeds.}
\label{tab:ablation}
\small
\renewcommand{\arraystretch}{1.2}
\begin{tabular}{l|cc|cc|cc}
\hline
\textbf{Method} & \textbf{Code F1 Mean} & \textbf{Std} & \textbf{Subcode F1 Mean} & \textbf{Std} & \textbf{Combo F1 Mean} & \textbf{Std} \\
\hline
PV-BERT-large & 78.62 & 1.30 & 76.21 & 1.75 & 73.94 & 1.75 \\
PV-BERT-large + author & 79.21 & 0.66 & 77.01 & 0.59 & 75.33 & 0.86 \\
PV-BERT-large + author + topic & \textbf{82.25} & 0.33 & \textbf{79.84} & 0.19 & \textbf{77.87} & 0.17 \\
\hline
\end{tabular}
\end{table*}

\section{Discussion}

\subsection{Clinical Implications}

Patient-generated communication data in general, particularly, patient-provider messages contain diverse expressions of patient voice, yet their unstructured nature makes large scale interpretation difficult. By structuring Codes, Subcodes, and Combo level patterns, PVminer enables more timely and data informed clinical workflows.

At the clinical level, identifying information giving and information seeking behaviors supports continuous monitoring between visits. Subcodes related to symptoms, drug questions, scheduling, and care coordination can signal emerging concerns that warrant prompt follow up. Automated recognition of these patterns may help care teams triage messages, prioritize high risk cases, and reduce delays in managing side effects or treatment barriers.

PVminer also captures relationship oriented behaviors that are largely absent from structured electronic health record fields. Categories such as PartnershipPatient alignment or PartnershipProvider maintain Communication reflect rapport, engagement, and shared understanding. Detecting these cues in text may help clinicians assess communication quality and identify patients who may need additional support.

In addition, recognizing SDOH related content such as EconomicStability or NeighborhoodAndBuilt Environment reveals broader contextual challenges that influence adherence and health outcomes. Integrating these signals into clinical workflows may support earlier identification of unmet needs and facilitate timely referral to support services.

Overall, PVminer offers a scalable foundation for improved triage, quality monitoring, and patient engagement in digital communication settings, with potential applications in message routing, longitudinal tracking, and decision support.

\subsection{Comparative Analysis}

To evaluate PVminer, we employ two domain-specific pre-trained encoders, PV-BERT-base and PV-BERT-large. Both encoders are integrated into the complete PVminer framework, in which topic keywords derived from the BERTopic model are appended to each input message during fine-tuning. Because topic-aware augmentation is a core component of PVminer, we do not separately evaluate standalone PV-BERT variants without topic enrichment. Accordingly, all reported results reflect the full PVminer design.

Given the complexity of multi-label communication coding and the inherent ambiguity of patient-authored messages, the achieved performance levels indicate that PVminer provides a stable and effective framework for structured patient voice extraction. The consistently strong results observed across diverse pre-trained encoders further demonstrate that PVminer generalizes well regardless of encoder origin, including general-domain, biomedical, and scientific models.

From the Code-level classification results, two key findings emerge. First, PVminer consistently achieves strong performance across all evaluated encoders, confirming its effectiveness as a unified framework for interpreting patient voice. Second, domain-specific pre-training on patient--provider communication data yields the most accurate and reliable representations for capturing high-level communicative intent in secure messages.

From the Subcode-level classification results, because all models operate within the same PVminer pipeline, two complementary conclusions can be drawn. First, PVminer maintains high absolute performance on a challenging multi-label task, indicating that topic-augmented and domain-aware modeling is effective for fine-grained patient voice extraction. Second, domain-specific pre-training on patient--provider messages produces the most accurate and reliable representations for distinguishing subcode categories. The particularly strong performance of PV-BERT-base underscores the value of aligning pre-training corpora with the linguistic characteristics of patient-generated text.

From the Combo-level classification results, this task proves to be the most challenging of the three, as it requires modeling interactions across multiple communicative dimensions and capturing subtle dependencies within composite labels. As expected, all models exhibit moderately lower performance compared to Code- and Subcode-level prediction. Nevertheless, because all encoders operate within the same PVminer pipeline, two important findings remain evident. First, PVminer achieves consistently strong performance on the most complex task in this study, with all models exceeding 73\% F1 despite the expanded label space and multi-label dependencies. Second, domain-specific pre-training on large-scale patient-generated text yields the most accurate composite label representations, confirming that PV-BERT encoders are better aligned with the linguistic and structural characteristics of patient-authored communication. Collectively, these results demonstrate that PVminer provides a robust and scalable framework for modeling the intertwined communicative and contextual signals that constitute the patient voice.

\subsection{PV Detection via PV-BERT-large}

From the Code-level results, PVminer demonstrates strong effectiveness in identifying structured, frequent, and lexically salient communication behaviors in patient--provider messages. These results indicate that the framework reliably captures high-level patient voice categories when linguistic cues are explicit and consistently expressed. At the same time, categories that are sparse, semantically abstract, or heavily context-dependent remain more challenging.

From the Subcode-level results, the PV-BERT-large encoder within the PVminer framework captures a broad range of fine-grained communicative behaviors, particularly those that are lexically grounded or structurally consistent across messages. However, subcodes that are infrequent, semantically subtle, or highly context-dependent continue to present difficulties, reflecting both data sparsity and the nuanced nature of certain communicative intents.

From the Combo-level results, the PV-BERT-large model demonstrates strong capacity to identify complex, multi-dimensional communicative behaviors when these combinations are frequent or expressed through stable lexical patterns. In contrast, low-resource, conceptually subtle, or variably expressed Code-Subcode combinations remain difficult to detect, highlighting the increased complexity introduced by composite labels.

\subsection{Challenges in Detecting Implicit and Structurally Complex Patient Voice Expressions}

Despite strong overall performance across Code-, Subcode-, and Combo-level tasks, PVminer remains challenged by patient voice categories that are infrequent, weakly lexicalized, or primarily conveyed through implicit meaning rather than explicit surface cues. In particular, Shared Decision-Making behaviors and certain Social Determinants of Health (SDOH) categories, most notably Neighborhood and Built Environment, exhibit consistently lower performance compared to more frequent and lexically salient categories.

Shared Decision-Making expressions, especially from the provider side, often require the model to infer intent from contextual negotiation, implied agreement, or multi-turn reasoning, rather than from explicit decision-oriented language. Collaborative decision processes may be embedded within clinical explanations or recommendations without overt markers such as ``decide,'' ``approve,'' or ``choose.'' In addition, these categories are relatively sparse in the annotated dataset, limiting the model’s ability to learn stable representations. 

Similarly, SDOH categories related to Neighborhood and Built Environment are challenging because relevant information is often indirectly referenced (e.g., transportation barriers, housing instability, or environmental constraints) rather than explicitly named. Such signals may be implied through descriptions of missed appointments, delayed care, or logistical difficulties, requiring broader contextual inference beyond keyword-based recognition. Moreover, these categories represent high-level constructs that implicitly encompass multiple underlying factors. These sub-sub domains that are not explicitly represented in the current annotation schema.

These challenges point to important directions for future work. Improving performance on implicit and structurally complex patient voice categories will likely require (1) expanding annotated data for low-frequency labels, (2) incorporating hierarchical or label-aware modeling strategies that better reflect nested decision-making and contextual structures, and (3) modeling conversational context across message threads to capture meaning that unfolds over multiple turns. Addressing these limitations is essential for advancing reliable extraction of nuanced patient voice elements that are critical for understanding shared decision-making processes and lived environmental constraints in patient-provider communication.

\subsection{Limitations and Future Directions}

This work is limited by the size and imbalance of the annotated dataset. Privacy constraints and the cost of expert annotation restrict access to richly labeled patient generated text. While PVminer benefits from large scale unlabeled pre-training, expanding the annotated corpus will be important for improving coverage of infrequent communicative behaviors.

Severe class imbalance further challenges learning. Because the task is multi label and multi class, simple up sampling is not feasible. Future work should develop augmentation strategies that respect the structured label space and maintain semantic fidelity. Adversarial augmentation represents a promising research direction, as well-constructed adversarial samples can maintain realistic data variations while substantially enhancing training sample diversity \cite{ma2022adaptive, ma2022regularization, ma2023improving, ma2023increasing, ma2023towards, ma2024general, ma2023towardsl}. This makes the resulting models more robust to unavoidable data noise while preserving or improving their generalization ability .


Performance is also reduced for rare categories that are underrepresented and often expressed indirectly, such as shared decision-making and Neighborhood and Built Environment. These categories rely on implied meaning rather than explicit lexical cues, making them more difficult to detect reliably. Future work will focus on enhancing the structure and design of PVMiner methods to better capture the nuances of such categories, which are grounded in the dynamics of how patients communicate their needs, social barriers, and involvement in navigating their plan of care. Expanding annotated examples, applying targeted oversampling strategies, and incorporating affect-aware or context-sensitive modeling approaches may further improve the detection of these Subcodes.

Finally, the current system treats each message independently. Many patient voice expressions unfold across multiple turns, and incorporating conversational context may improve recognition of shared decision making, longitudinal concerns, or evolving symptoms. Future research should explore hierarchical encoders, conversation level representations, and temporal modeling.

Addressing these limitations will strengthen the reliability of automated patient voice analysis and improve integration into clinical workflows.

\section{Conclusion}

This study formulates patient voice detection in secure patient-provider messages as a multi-label, multi-class classification task over Codes, Subcodes, and Combo-level labels. In doing so, it defines a structured prediction problem that captures overlapping communicative and social dimensions of patient-generated text, rather than constraining messages to a single dominant category.

We present PVminer, a domain-adapted NLP framework for structuring patient voice in secure patient-provider communication. The framework integrates three components: pre-training of patient-specific BERT encoders on large-scale patient-generated messages, topic modeling for thematic augmentation, and multi-label prediction of hierarchical labels. Within this framework, the PV-BERT-base and PV-BERT-large encoders achieve higher performance than general-domain and biomedical pre-trained models across all classification tasks, demonstrating the value of pre-training directly on patient-authored text.

The ablation study further shows that each component of PVminer contributes meaningfully to overall performance. Incorporating author information yields measurable gains, and adding topic-based augmentation provides the largest improvement across Code, Subcode, and Combo classification. These findings confirm that PVminer effectively combines linguistic, contextual, and thematic cues to enhance the structured interpretation of patient-generated messages.

By transforming unstructured text into structured communicative and social signals, PVminer offers foundational infrastructure for downstream applications such as triage support, quality monitoring, and analyses of how patients express needs, preferences, and barriers in digital communication. At the same time, the work highlights persistent challenges related to data imbalance, detection of emotionally nuanced categories, and the absence of conversation context when classifying single messages.

Future work should expand annotated resources, explore label-aware augmentation and adversarial training strategies, and develop context-sensitive modeling of patient-provider exchanges over time. Advancing these directions will further strengthen the ability of PVminer-based systems to interpret patient voice and support more patient-centered clinical decision making.

\section{Competing interests}
No competing interest is declared.

\section{Funding Statement}

This work was supported by the Patient-Centered Outcomes Research Institute (PCORI) under Award No. ME-2023C2-31367 (to S.F.).

\section{Author Contributions Statement}

S.F. conceptualized the study, designed the methodology and data analysis plan, and led the writing and revision of the manuscript. L.M. conducted the experiments and the analysis, contributed to the interpretation of the results, and co-wrote the manuscript. Y.W., S.T., G.P., and A.K. assisted with conducting the experiments. A.H., S.L., and A.R. provided domain expertise in patient care and supported the accuracy and integrity of the content. All authors reviewed and approved the final manuscript.

\section{Data Availability Statement}

The data analyzed in this study consist of de-identified patient–provider secure messages and associated annotations derived from clinical systems. Due to privacy, ethical, and institutional restrictions, these data are not publicly available. Access to the data may be considered upon reasonable request to the corresponding author and with appropriate institutional review board (IRB) approval and data use agreements in place.


\bibliographystyle{plain}
\bibliography{reference}

\clearpage

\section{Appendix: A. Code and Subcode Definitions}\label{sec_codebook}



The annotation schema consists of eight major Codes, each representing a distinct communicative or social construct within the patient voice. Each major code has a corresponding set of \textit{Subcodes} that capture more specific communicative intents. Below are concise summaries of  these key categories:

\vspace{0.5em}
\noindent\textbf{1. Social Determinants of Health (SDOH).}  
Refers to contextual, social, and environmental conditions influencing an individual's health and well-being.  
\textit{Subcodes:}
\begin{itemize}
    \item \textbf{Economic Stability:} Financial security and resources to afford healthcare, housing, food, and other necessities. 
    \item \textbf{Education Access and Quality:} Availability and quality of educational opportunities and the ability to obtain and apply health knowledge.
    \item \textbf{Healthcare Access and Quality:} Ability to obtain necessary health services and the standard of care received. 
    \item \textbf{Neighborhood and Built Environment:} Physical and social surroundings such as housing, transportation, safety, and environmental quality. 
    \item \textbf{Social and Community Context:} Social relationships, interactions, and community-level conditions affecting health outcomes.
\end{itemize}

\noindent\textbf{2. Partnership from the Patient Side.}  
Captures patients’ active engagement and collaboration in their healthcare through communication and participation.  
\textit{Subcodes:}
\begin{itemize}
    \item \textbf{Active Participation/Involvement:} Patient provides information, asks questions, or contributes to decision-making.
    \item \textbf{Express Opinions:} Sharing thoughts, perspectives, or feedback regarding care and treatment. 
    \item \textbf{Signoff:} Courteous closure that marks completion of a message. 
    \item \textbf{State Preferences:} Communicating personal values or desired treatment approaches. 
    \item \textbf{Alignment:} Building mutual understanding and trust-based relationships with the provider. 
    \item \textbf{Appreciation/Gratitude:} Expressing thanks or acknowledgment of provider support. 
    \item \textbf{Connection:} Non-clinical or relational communication that strengthens rapport. 
    \item \textbf{Salutation:} Greeting or addressing the provider by name or title. 
    \item \textbf{Clinical Care:} Describing symptoms, treatments, or medical experiences from the patient’s perspective. 
    \item \textbf{Build Trust:} Expressing confidence in the provider’s competence and integrity. 
\end{itemize}

\noindent\textbf{3. Partnership from the Provider Side.}  
Represents providers’ behaviors promoting collaboration, equality, and patient empowerment.  
\textit{Subcodes:}
\begin{itemize}
    \item \textbf{Invite Collaboration:} Encouraging patients to participate in care or decision-making. 
    \item \textbf{Request for Opinion:} Seeking patient input or feedback on treatment decisions. 
    \item \textbf{Checking Understanding/Clarification:} Confirming patient comprehension of information. 
    \item \textbf{Appreciation/Gratitude:} Expressing thanks or acknowledgment toward the patient. 
    \item \textbf{Signoff:} Courteous ending signaling completion of communication. 
    \item \textbf{Acknowledge Patient Expertise/Knowledge:} Recognizing the patient’s experiential knowledge. 
    \item \textbf{Maintain Communication:} Informing the patient about next steps or updates in care. 
    \item \textbf{Alignment:} Ensuring mutual understanding and shared goals between provider and patient. 
    \item \textbf{Connection:} Relational remarks that enhance rapport beyond medical context. 
    \item \textbf{Salutation:} Greeting or addressing the patient respectfully. 
    \item \textbf{Clinical Care:} Delivering or planning medical treatment, diagnosis, or procedures. 
    \item \textbf{Build Trust:} Establishing confidence through transparency and reliability. 
\end{itemize}

\noindent\textbf{4. Shared Decision-Making from the Patient Side.}  
Denotes patients’ participation in exploring and approving treatment options aligned with personal values.  
\textit{Subcodes:}
\begin{itemize}
    \item \textbf{Explore Options:} Asking about or discussing possible treatment choices. 
    \item \textbf{Seeking Approval:} Requesting permission or confirmation from the provider. 
    \item \textbf{Approval of Decision:} Expressing agreement or affirmation of a chosen plan. 
\end{itemize}

\noindent\textbf{5. Shared Decision-Making from the Provider Side.}  
Captures providers’ facilitation of collaborative decision-making through two-way dialogue.  
\textit{Subcodes:}
\begin{itemize}
    \item \textbf{Share Options:} Presenting available alternatives and explaining benefits or risks. 
    \item \textbf{Summarize and Confirm Understanding:} Recapping and verifying patient comprehension of discussed options. 
    \item \textbf{Make Decision:} Finalizing a treatment plan jointly with the patient. 
    \item \textbf{Approval of Decision/Reinforcement:} Providing validation and encouragement after decision-making. 
\end{itemize}

\noindent\textbf{6. Socioemotional Behavior.}  
Encompasses empathy, reassurance, appreciation, and other affective or relational expressions demonstrating emotional understanding and support. No Subcodes are defined.

\noindent\textbf{7. Care Coordination Provider.}  
Refers to the provider’s organization and integration of healthcare services to ensure effective and timely patient care. No Subcodes are defined.

\noindent\textbf{8. Care Coordination Patient.}  
Represents the patient’s communication efforts to coordinate care among multiple providers or services. No Subcodes are defined.

These hierarchical categories enable PVminer to capture both the content and the relational intent embedded in patient discourse, supporting downstream analyses that link language use to social context, decision-making, and health outcomes.

\end{document}